\documentclass[10pt,twocolumn,letterpaper]{article}

\usepackage[pagenumbers]{cvpr} 

\usepackage{graphicx}
\usepackage{amsmath}
\usepackage{booktabs}
\usepackage{amssymb}
\usepackage{latexsym}
\usepackage{framed,multirow}

\usepackage{hhline}
\usepackage{array}
\usepackage{wrapfig}
\usepackage{floatrow, comment, enumerate}
\usepackage{color}
\usepackage{algorithmic}
\usepackage[linesnumbered,ruled,vlined]{algorithm2e}
\usepackage[utf8]{inputenc}
\usepackage{paralist}

\DeclareUnicodeCharacter{2212}{-}

\usepackage{url}
\usepackage{xcolor}
\definecolor{newcolor}{rgb}{.8,.349,.1}

\let\oldnl\nl
\newcommand{\nonl}{\renewcommand{\nl}{\let\nl\oldnl}}

\usepackage[pagebackref=true,breaklinks=true,colorlinks,bookmarks=false,citecolor=blue,linkcolor=blue]{hyperref}

\usepackage[capitalize]{cleveref}
\crefname{section}{Sec.}{Secs.}
\Crefname{section}{Section}{Sections}
\Crefname{table}{Table}{Tables}
\crefname{table}{Tab.}{Tabs.}

\def\cvprPaperID{3139} 
\def\confName{CVPR}
\def\confYear{2023}

\begin{document}

\title{Multimodal Prompting with Missing Modalities for Visual Recognition}

\author{%
  Yi-Lun Lee\textsuperscript{$\dagger$} \quad Yi-Hsuan Tsai\textsuperscript{$\ddagger$} \quad Wei-Chen Chiu\textsuperscript{$\dagger$} \quad Chen-Yu Lee\textsuperscript{$\ddagger$} \\
  \textsuperscript{$\dagger$}National Yang Ming
Chiao Tung University \quad \textsuperscript{$\ddagger$}Google\\  
  \small{\texttt{\{yllee10727, walon\}@cs.nctu.edu.tw}}, \;  
  \small{\texttt{\{yhtsai, chenyulee\}@google.com}}
}

\maketitle

\begin{abstract}
In this paper, we tackle two challenges in multimodal learning for visual recognition: 1) when missing-modality occurs either during training or testing in real-world situations; and 2) when the computation resources are not available to finetune on heavy transformer models.
To this end, we propose to utilize prompt learning and mitigate the above two challenges together. Specifically, our modality-missing-aware prompts can be plugged into multimodal transformers to handle general missing-modality cases, while only requiring less than $1\%$ learnable parameters compared to training the entire model.
We further explore the effect of different prompt configurations and analyze the robustness to missing modality.
Extensive experiments are conducted to show the effectiveness of our prompt learning framework that improves the performance under various missing-modality cases, while alleviating the requirement of heavy model re-training. Code is available.\footnote{\href{https://github.com/YiLunLee/Missing_aware_prompts}{https://github.com/YiLunLee/missing\_aware\_prompts}}
\end{abstract}
\section{Introduction}
Our observation perceived in daily life is typically mulitmodal, such as visual, linguistic, and acoustic signals, thus modeling and coordinating multimodal information is of great interest and has broad application potentials.
Recently, multimodal transformers\cite{kim2021vilt,li2021align,ma2022multimodal,pham2019found,wang2020transmodality} emerge as the pre-trained backbone models in several multimodal downstream tasks, including genre classification~\cite{ma2022multimodal}, multimodal sentiment analysis~\cite{pham2019found,wang2020transmodality}, and cross-modal retrieval~\cite{kim2021vilt,li2021align,lee2022formnet,saito2023pic2word}, etc. 
Though providing promising performance and generalization ability on various tasks, there are still challenges for multimodal transformers being applied in practical scenarios: 1) how to efficiently adapt the multimodal transformers without using heavy computation resource to finetune the entire model? 2) how to ensure the robustness when there are missing modalities, e.g., incomplete training data or observations in testing?
\begin{figure}[!t]
    \centering
    \includegraphics[width=\textwidth]{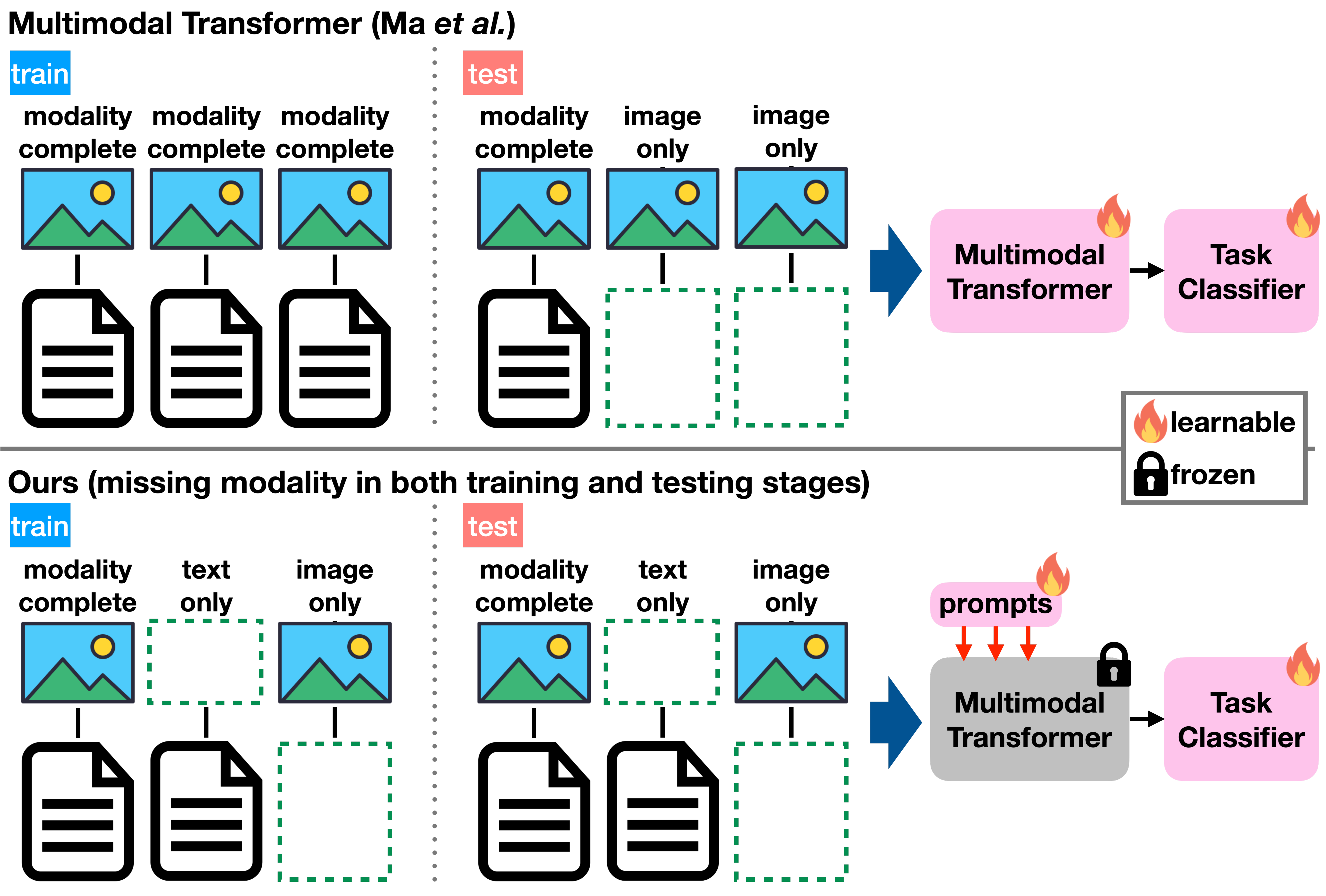}
    \vspace{-2em}
    \caption{
    Illustration of missing-modality scenarios in training multimodal transformers. Prior work~\cite{ma2022multimodal} investigates the robustness of multimodal transformers to modality-incomplete test data, with the requirement to finetune the entire model using modality-complete training data. In contrast, our work studies a more general scenario where various modality-missing cases would occur differently not only for each data sample but also learning phases (training, testing, or both), and we adopt prompt learning to adapt the pre-trained transformer for downstream tasks without requiring heavy computations on finetuning the entire model.
   }
\label{fig:teaser}
\end{figure}

Most multimodal transformer-based methods have a common assumption on the data completeness, which may not hold in practice due to the privacy, device, or security constraints. Thus, the performance may degrade when the data is modality-incomplete (regardless of training or testing). On the other hand, transformers pretrained on large-scale datasets are frequently adopted as backbone and finetuned for addressing various downstream tasks, thanks to the strong generalizability of transformers. 
However, as the model size of transformers increases (e.g., up to billions of parameters~\cite{brown2020language,raffel2020exploring,rae2021scaling}), finetuning becomes significantly expensive (e.g., up to millions of A100-GPU-hours~\cite{scao2022language}) and is even not feasible for practitioners due to the limited computation resources in most real-world applications. 
In addition, finetuning a transformer on relatively small-scale target datasets can result in restricted generalizability~\cite{kejriwal2020fine,ju2022robust} and stability~\cite{mosbach2020stability}, thus hindering it from being reused for further learning with new data or in other tasks.

This motivates us to design a method that allows multimodal transformers to alleviate these two real-world challenges.
One pioneer work~\cite{ma2022multimodal} investigates the sensitivity of vision-language transformers against the presence of modal-incomplete test data (i.e., either texts or images are missing).
However, they only consider the case of missing a specific modality for all the data samples, while in real-world scenarios the missing modality for each input data could not be known in advance.
Moreover, \cite{ma2022multimodal} introduces additional task tokens to handle different missing-modal scenarios (e.g., text-only token when missing visual modality) and requires to optimize cross-modal features in the model.
Hence finetuning the entire transformer becomes inevitable, leading to significant computation expense.

In this paper, we study multimodal transformers under a more general modality-incomplete scenario, where various missing-modality cases may occur in any data samples, e.g., there can be both text-only and image-only data during training or testing.
In particular, we also focus on alleviating the requirement of finetuning the entire transformers. To this end, we propose a framework stemmed from prompt learning techniques for addressing the aforementioned challenges. 
Basically, prompt learning methods~\cite{brown2020language,lester2021power,li2021prefix,jia2022visual,bahng2022visual,tsimpoukelli2021multimodal,zhou2022learning}
emerge recently as efficient and effective solutions for adapting pre-trained transformers to the target domain via only training very few parameters (i.e., prompts), and achieve comparable performance with finetuning the whole heavy model. 
As motivated by~\cite{saito2022prefix} which shows that prompts are good indicators for different distributions of input, we propose to regard different situations of missing modalities as different types of input and adopt the learnable prompts to mitigate the performance drop caused by missing modality. As a result, the size of our learnable prompts can be less than 1\% of the entire transformer, and thus the computation becomes more affordable compared to holistic finetuning. The key differences between our work and \cite{ma2022multimodal} are illustrated in Figure~\ref{fig:teaser}. 

In order to further explore the prompt designs for multimodal transformers to tackle the general modality-incomplete scenario, we investigate two designs of integrating our missing-aware prompts\footnote{In this paper, we use ``missing-aware prompts'' and ``modality-missing-aware prompts'' interchangeably.} into pre-trained multimodal transformers: 1) input-level, and 2) attention-level prompt learning.
We find that, the location of attaching prompts to transformers is crucial for the studied missing-modality cases in this paper, which also aligns the findings in \cite{wang2022dualprompt}, though under a different problem setting.

We conduct experiments to explore different prompt configurations and have observations of the impact on the length and location of prompts:
1) As the number of prompting layers increases, the model performs better intuitively but it is not the most important factor;
2) Attaching prompts to the layers near the data input achieves better performance;
3) The prompts' length has slight impact on model performance for attention-level prompts but may influence input-level prompts more on certain datasets. 
Moreover, we show extensive results to validate the effectiveness of adopting our prompting framework to alleviate the missing-modality issue under various cases, while reducing the learnable parameters to less than $1\%$ compared to the entire model.
Our main contributions are as follows:
\begin{compactitem}
    \item We introduce a general scenario for multimodal learning, where the missing modality may occur differently for each data sample, either in training or testing phase.

    \item We propose to use missing-aware prompts to tackle the missing modality situations, while only requiring less than 1\% parameters to adapt pre-trained models, thus avoiding finetuning heavy transformers. 
    
    \item 
    We further study two designs of attaching prompts onto different locations of a pretrained transformer, input-level and attention-level prompting, where the input-level prompting is generally a better choice but the attention-level one can be less sensitive to certain dataset settings.
    
\end{compactitem}

\begin{figure*}[!t]
    \centering
    \includegraphics[width=.95\textwidth]{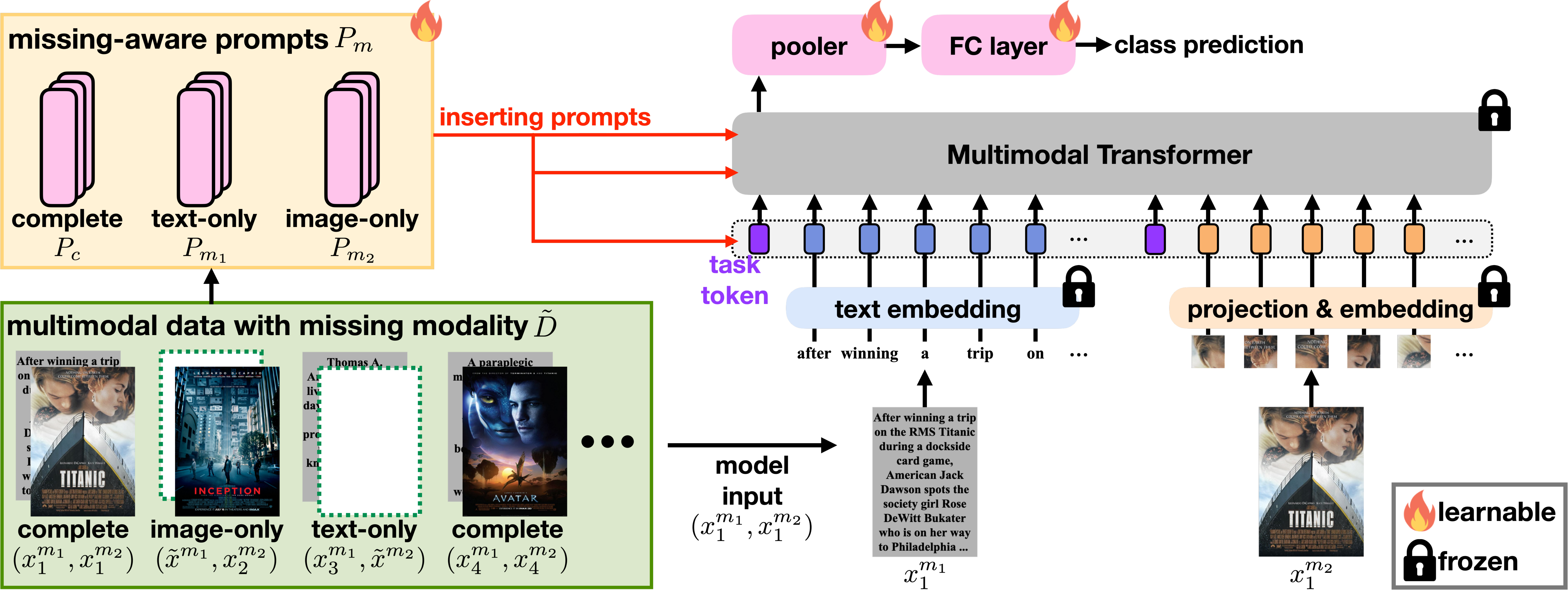}
    \vspace{-.5em}
    \caption{The overview of our proposed prompt-based multimodal framework. We first select the missing-aware prompts $P_m$ according to the missing case (e.g., complete, text-only, image-only in vision-language tasks) of the multimodal inputs ($x^{m_1}_i,x^{m_2}_i)$, in which the dummy inputs $\{\Tilde{x}^{m_1}$, $\Tilde{x}^{m_2}\}$ respectively for text and image are adopted for the corresponding missing modality. Then we attach missing-aware prompts into multiple MSA layers via different prompting approaches (see Figure~\ref{fig:prompt_function} and Section~\ref{sec:method_prompt_func}). We select the text-related task token of the multimodal transformer as our final output features, and feed them to the pooler layer and fully-connected (FC) layers for class predictions. Note that only the pink-shaded blocks require to be trained while the others are frozen.
   }
\label{fig:model}
\end{figure*}

\section{Related Work}
\paragraph{Missing-Modality for Multimodal Learning.} 
Multimodal learning methods leverage the complementary property from different modalities (e.g., images, texts, or audio) for learning to describe a common concept. 
Recently, multimodal transformers emerge as unified models that process inputs from different modalities and fuse them via token concatenation without modality-specific feature extractors. Such transformer-based models are widely applied in various mutlimodal tasks~\cite{kim2021vilt,li2021align,gabeur2020multi,botach2022end}. 
However, most multimodal learning methods require the completeness of modality, which may not be the real-world case. Once one of the modalities is missing, the multimodal fusion becomes unreachable and the model may predict inaccurately \cite{ma2022multimodal}.

To this end, recent works~\cite{ma2021smil,zeng2022tag,ma2022multimodal,zhao2021missingimagenet} explore to build multimodal models which are robust to data with missing modalities. SMIL~\cite{ma2021smil} is proposed to estimate the latent features of the modality-incomplete data via Bayesian Meta-Learning. Zeng~\etal~\cite{zeng2022tag} propose a tag encoding module to assist the transformer's encoder learning with different missing modalities. MMIN~\cite{zhao2021missingimagenet} predicts the representation of any missing modality given other available modalities via learning a joint multimodal representations. Ma~\etal~\cite{ma2022multimodal} investigate the robustness of multimodal transformers to missing modalities and improve it via multi-task optimization to achieve better fusion strategies respectively for different missing-modality cases.
In contrast, this paper conducts a more thorough study on multimodal transformer's robustness where various modality-missing would occur for any data samples and anywhere in learning phases, particularly focusing on reducing the computation of model finetuning.

\vspace{-2mm}
\paragraph{Prompt Learning.} 
Prompt learning emerges as an effective transfer learning technique in nature language processing, which adopts a ``prompt'' to modify the input text for instructing the pre-trained model for downstream tasks. 
For instance, with manually-chosen prompts, the pre-trained language models~\cite{brown2020language} show a strong generalizability to downstream tasks in few-shot or zero-shot manner. 
Instead of relying on human involvement to discover the proper prompts for adapting transformers into new tasks, prompt-tuning~\cite{lester2021power} and prefix-tuning~\cite{li2021prefix} are proposed to automate the prompt learning in continuous space.   

Recently, prompts are also introduced into computer vision tasks~\cite{jia2022visual,bahng2022visual,wang2022learning, wang2022dualprompt} and multimodal learning tasks~\cite{yang2022prompting,tsimpoukelli2021multimodal,zhou2022learning,khattak2022maple, liang2022modular, zhao2022memobert}. 
The visual prompts~\cite{jia2022visual,bahng2022visual} are applied to the vision transformers and adapt the large-scale pre-trained model to downstream vision tasks via tuning very few learnable parameters. L2P~\cite{wang2022learning} and DualPrompt~\cite{wang2022dualprompt} further adopt the prompts to learn different task information conditionally in continual learning.
CoOp~\cite{zhou2022learning} models the context in prompts with learnable vectors in continuous space while the parameters of the entire CLIP-like pre-trained model are kept fixed. Frozen~\cite{ma2022multimodal} encodes the image as a sequence of continuous embeddings to serve as a prefix prompt to instruct the pre-trained frozen language models in generating the appropriate caption by a multimodal few-shot learning manner.
MaPLe~\cite{khattak2022maple} further applies prompts in both vision and language encoders to improve the alignment between vision and language representations.
Besides, PromptFuse~\cite{liang2022modular} utilized prompts to learn alignment among different modalities for parameter-efficient adaptation on downstream tasks.
These works explore the great adaptation ability of prompt learning to different tasks with different input domains. This motivates us to integrate the prompt learning technique into multimodal learning under a general missing-modality scenario via regarding different missing cases as different learning tasks.

\section{Proposed Method}
\label{sec:method}

\subsection{Overall Framework}

In this paper, we focus on multimodal learning with missing modalities in general situations. We assume that there are several missing-modality cases, e.g., missing one modality or missing more modalities, to represent the more realistic scenario of multimodal learning in the real world.
Note that, the missing case during training can be also different from the one in testing.
In addition, as the pretrained transformers become larger and untrainable with limited computation resources, it is crucial to develop the method without the need of finetuning the entire pretrained model.

\vspace{-4mm}
\paragraph{Problem Definition.}
To be the simplest but without loss of generality, we consider a multimodal dataset consisting of $M=2$ modalities $m_1$ and $m_2$ (e.g., image and text). Given a multimodal dataset $D = \{D^c, D^{m_1}, D^{m_2}\}$, we denote $D^c = \{x^{m_1}_i, x^{m_2}_i,y_i\}$ as the modality-complete subset, while $D^{m_1} = \{x^{m_1}_j,y_j\}$ and $D^{m_2} = \{x^{m_2}_k, y_k\}$ are denoted respectively as the modality-incomplete subsets (e.g., text-only and image-only) where one modality is missing. 
As shown in Figure~\ref{fig:model}, the training data may contain data samples with different missing cases including complete data $D^c$, text-only data $D^{m_1}$, and image-only data $D^{m_2}$.

To preserve the format of multimodal inputs, we simply assign dummy inputs $\Tilde{x}^{m_1}$, $\Tilde{x}^{m_2}$ (e.g., empty string/pixel for texts/images) to the missing-modality data and obtain $\Tilde{D}^{m_1} = \{x^{m_1}_j,\Tilde{x}^{m_2}_j,y_j\}, \Tilde{D}^{m_2} = \{\Tilde{x}^{m_1}, x^{m_2}_k, y_k\}$. Therefore, the multimodal data with missing modality can be reformed as $\Tilde{D} = \{D^c, \Tilde{D}^{m_1}, \Tilde{D}^{m_2}\}$.

For simplicity, we follow \cite{ma2022multimodal} to adopt the multimodal transformer ViLT~\cite{kim2021vilt} as our backbone model, which is pre-trained on large-scale vision and language datasets. Note that the backbone model is untrainable in our scenario due to the limitation of computation resources.
In order to tackle the missing modality, we propose \textbf{missing-aware prompts} to instruct the pretrained model's prediction with different input cases. These prompts are assigned according to the missing case of input data and attached to multiple blocks of the multimodal transformer. With the assumption of untrainable pretrained models, the only trainable parameters are the missing-aware prompts, pooler layer, and fully-connected layers for learning the multimodal classifier.

\subsection{Prompt Learning for Missing Modalities}
Prompt-based learning first emerges as the efficient method in natural language processing (NLP) for transfer learning without finetuning the whole pretrained model.
In general, prompts are prepended to the input for instructing the model prediction.
With a similar motivation, we propose missing-aware prompts to instruct the pretrained transformer, conditioned on different input cases of missing modality.
To this end, we design the corresponding missing-aware prompts for each missing-modality case.
As shown in Figure~\ref{fig:model}, we first assign $M^2-1$ prompts for $M$ modality tasks (e.g., $3$ missing-aware prompts for the vision-language task), and prepend them to the input according to the type of missing modality. 

Given a pretrained multimodal transformer $f_{\theta}$ with $N$ consecutive MSA layers, we denote the input embedding features of the $i$-th MSA layer as $h^i \in \mathbb{R}^{L \times d}, i=1,2,...,N$ with input length $L$ and embedding dimension $d$. Note that $h^1$ is the output of modality-specific embedding functions that pre-process the inputs to token sequences (i.e., BERT tokenizer for the text modality and visual embedding~\cite{dosovitskiy2020vit} layers for the image modality). Then the missing-aware prompts $p^i_m \in \mathbb{R}^{L_p \times d}$ are attached to the $i$-th layer, where $L_p$ is the prompt length, $d$ is the embedding dimension, and $m \in \{c, m_1, m_2\}$ represents different missing-modality cases.
Finally, the missing-aware prompts are attached to the embedding features along with the input-length dimension to form extended features $h^i_p$:
\begin{equation}
h^i_p = f_{prompt}(p^i_m, h^i),
\label{eq:prompt}
\end{equation}
where $f_{prompt}$ defines the approach to attach prompts to the embedding features, and will be detailed in the next section.

\vspace{-4mm}
\paragraph{Overall Objective.}
For model training, we freeze all the parameters $f_{\theta}$ of the multimodal transformer except for the task-specific layers $f_{\theta_t}$ (i.e., pooler layer and fully-connected layer), in order to output corresponding predictions based on each visual perception task.
Moreover, we denote $\theta_p$ as the parameters of missing-aware prompts.
The overall objective with trainable parameters is defined as:
\begin{equation}
\label{equ:attenion_prompt} 
L = L_{task}(x^{m_1}_i, x^{m_2}_i; \theta_t, \theta_p),
\end{equation}
where $(x^{m_1}_i, x^{m_2}_i) \in \Tilde{D}$ is the multimodal input pair with missing-modality cases, and $L_{task}$ represents the task-specific multimodal objective, e.g., binary cross-entropy loss for movie genre classification.

\begin{figure}[!t]
    \centering
    \includegraphics[width=0.95\textwidth]{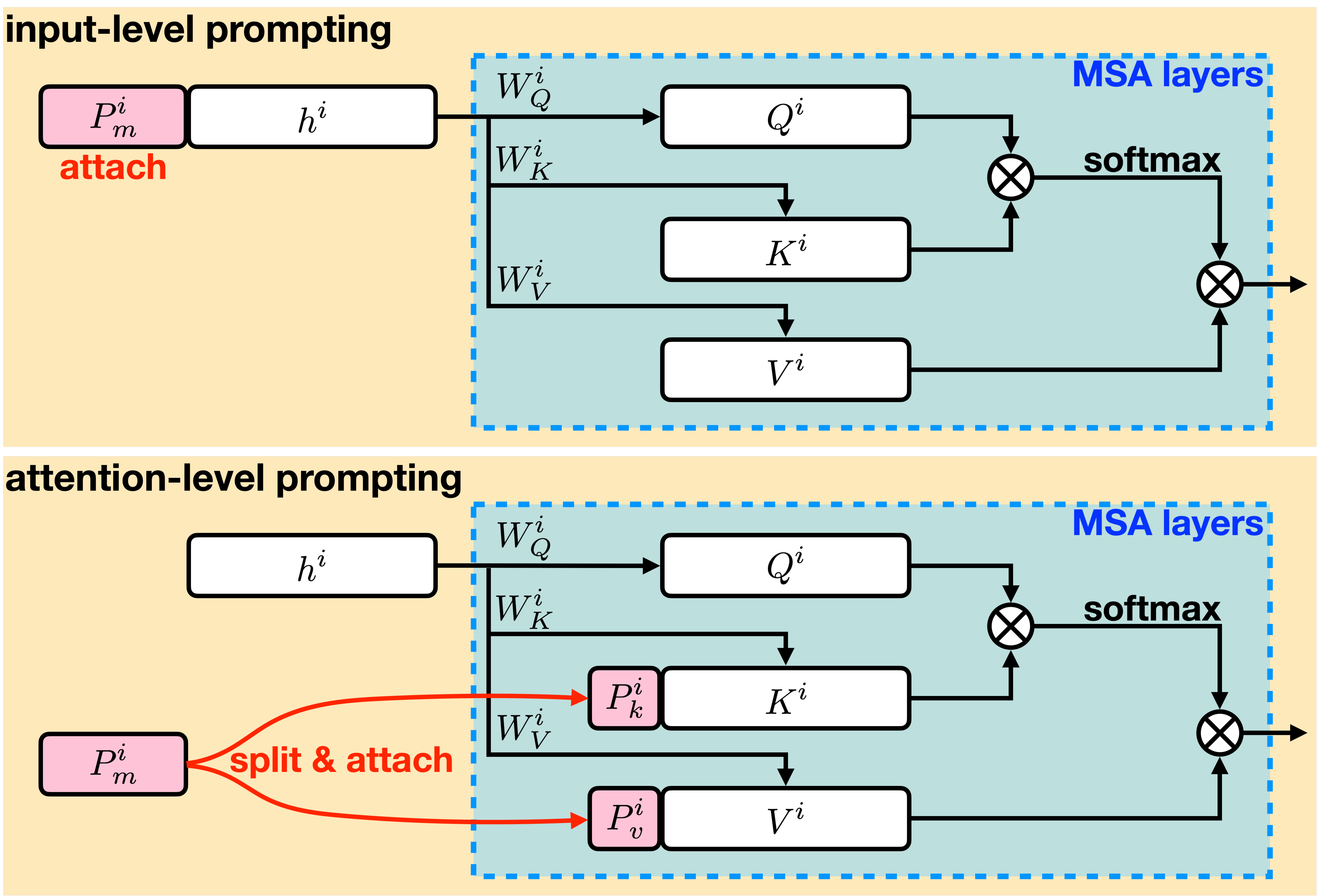}
    \vspace{-.5em}
    \caption{The illustration of two prompting approaches. The top block shows the input-level prompting method, which attaches the missing-aware prompts $p_m^i$ to the input of the $i$-th MSA layer. The bottom block shows the attention-level prompting method, which first splits the missing-aware prompts $p_m^i$ into two sub-prompts $p^i_k, p^i_v$ with the same length, and attaches them respectively to the key $K^i$ and value $V^i$ in the $i$-th MSA layer (see Section~\ref{sec:method_prompt_func}).
   }
\label{fig:prompt_function}
\end{figure}

\subsection{Prompt Design}
\label{sec:method_prompt_func}
In this section, we focus on the design of the $f_{prompt}$ function that attaches prompts to each input layer as in \eqref{eq:prompt}.
In general, most prompt-based methods typically add the prompts to the input sequence and instruct the model for downstream tasks. 
However, \cite{wang2022dualprompt} shows that the configuration of prompts and the position where prompts are added is crucial to prompt-based learning.
In our situation, since the input modality may be missing, studying the proper configuration to attach prompts is of great importance.
In Figure~\ref{fig:prompt_function}, we introduce two configurations of prompts: \textit{input-level prompting} and \textit{attention-level prompting}.

\vspace{-4mm}
\paragraph{Input-level Prompting.}
A common approach to attach the prompts is to prepend prompts into input sequences for each layer, as shown in the top of Figure~\ref{fig:prompt_function}. The prompt function can be written as:
\begin{equation}
\label{equ:input_prompt} 
\begin{aligned}
f^{input}_{prompt}(p^i_m, h^i) = [p^i_m;h^i],
\end{aligned}
\end{equation}
where $[\cdots;\cdots]$ represents the concatenation operation. Assume there are $N_p$ layers attaching the prompt parameters, the length of input/output sequence for each MSA layer would become larger as it goes deeper. For example, the length of sequence in the output of the last MSA layer with prompts would become $(N_pL_p+L)$.
In this way, the prompts for the current layer can interact with the prompt tokens inheriting from previous layers, and thus learn more effective instructions for the model prediction. 
However, we find that such increasing length of input sequence makes the input-level prompt learning sensitive to the dataset with different multimodal token lengths, which may be less favorable in certain multimodal downstream tasks. We discuss the details in Section~\ref{sec:performance_sensitivity}.

\vspace{-4mm}
\paragraph{Attention-level Prompting.}
Another prompting approach is to modify the inputs of the MSA layers~\cite{vaswani2017attention} with prompts. In the bottom of Figure~\ref{fig:prompt_function}, we split the prompt into two sub-prompts $p^i_k, p^i_v$ with the same sequence length $\frac{L_p}{2}$ and prepend them to the key and value vectors respectively. 
We denote the query, key and value for the MSA layer as:
\begin{equation}
\label{equ:MSA_proj} 
\begin{aligned}
Q^i = h^i W^i_Q; K^i = h^i W^i_K; V^i = h^i W^i_V,
\end{aligned}
\end{equation}
where $W^i_Q,W^i_K,W^i_V \in \mathbb{R}^{d\times d}$ is the projection weights for MSA layers. Then, we can define the prompt function for attention-level prompts as:
\begin{equation}
\label{equ:attenion_prompt} 
\begin{aligned}
&f^{attn}_{prompt}(p^i_m, h^i) = \textsc{Attention}^i(p^i_m,h^i),\\
&\textsc{Attention}^i = softmax(\frac{Q^i[p^i_k,K^i]^T}{\sqrt{d}})[p^i_v;V^i].
\end{aligned}
\end{equation}
The attention-level prompting provides another way to instruct the pretrained model from the perspective of the attention mechanism in transformers. As the prompts do not prepend to the query vector, the output sequence length remains the same as the input sequence.

\vspace{-4mm}
\paragraph{Multi-layer prompting and locations where to attach prompts.}
Intuitively, different layers of the multimodal transformer have different context of feature embedding~\cite{raghu2021vision}, and the effect of prompts for each layer may vary.
With the self-attention mechanism, the input tokens from different modalities are fused closely along with the transformer layers.
That is, the features of early layers may have more characteristics from different modalities than those of deeper layers that are well-fused to multimodal tokens with respect to the task objective. This motivates us to explore the most proper locations to attach missing-aware prompts.

Here, we introduce the multi-layer extension of prompts which can be denoted as $P_m = \{p^i_m\}_{i=start}^{end} \in \mathbb{R}^{N_p \times L_p \times d}$, where $p^i_m$ is the prmopt attaching to the input sequence (input-level) or MSA layer (attention-level) of the $i$-th layer in transformers, and $N_p = (end-start+1)$ is the total number of layers with prompts. Note that we simply assume that the chosen indices of MSA layers are continuous.
Instead of attaching prompts to either whole layers or only the first layer, we empirically find that early half of layers is the best location starting from the first layer ($start=0, end=\frac{N}{2}-1)$ with $N_p = \frac{N}{2}$.
More results and discussions are in Section \ref{sec:prompt_position}.

\section{Experimental Results}
\vspace{-2mm}
\paragraph{Datasets.} 
We follow the work~\cite{ma2022multimodal} to evaluate our methods on three multimodal downstream tasks:

$\bullet$~\textit{MM-IMDb}~\cite{arevalo2017mmimdb}  is a movie genre classification dataset with image and text modalities. As a movie might have several genres, the task is hence a multi-label classification that predicts the genre via using image, text, or both modalities. 

$\bullet$~\textit{UPMC Food-101}~\cite{wang2015food101} is the classification dataset including image and text, in which it consists of noisy image-text paired data collected from Google Image Search and has identical categories to the largest publicly available ETHZ Food-101 dataset~\cite{bossard2014food}.

$\bullet$~\textit{Hateful Memes}~\cite{kiela2020hateful} is a more challenging multimodal dataset that aims to identify the hate speech in memes via using image and text modalities. To prevent the model from relying on a single modality, it is constructed to make unimodal models more likely to fail via adding challenging samples (``benign confounders'') while the multimodal models are more likely to work better. 

\noindent\textbf{Metrics.}~ 
As these datasets focus on different classification tasks, we use corresponding proper metrics for each dataset.
For \textit{MM-IMDb}~\cite{arevalo2017mmimdb}, F1-Macro is adopted
to measure the multi-label classification performance;
For \textit{UPMC Food-101}~\cite{wang2015food101}, the metric is the classification accuracy;
For \textit{Hateful Memes}~\cite{kiela2020hateful}, we use Area Underthe Receiver Operating Characteristic Curve (AUROC).

\begin{table*}[!t]
    \centering
    \caption{Quantitative results on the MM-IMDB~\cite{arevalo2017mmimdb}, UPMC Food-101\cite{wang2015food101}, and Hateful Memes~\cite{kiela2020hateful} datasets with missing rate $\eta\% = 70\%$ under various modality-missing scenarios. \textbf{Bold} number indicates the best performance.
    } 
    \label{table:qualitative}
    \resizebox{0.9\linewidth}!{
    \begin{tabular}{c|c|cccc|ccc}
    \hline
    \multirow{2}{*}{Datasets} & 
    \multirow{2}{*}{\shortstack{Missing \\ rate $\eta$}} & 
    \multicolumn{2}{c}{Training} &
    \multicolumn{2}{c|}{Testing} & 
    \multirow{2}{*}{Baseline\cite{kim2021vilt}} &
    \multirow{2}{*}{\shortstack{Attention-level \\ prompts (Ours)}} &
    \multirow{2}{*}{\shortstack{Input-level \\ prompts (Ours)}}
    \\ 
    &&Image &Text &Image &Text
    \\ \hline
    \multirow{3}{*}{\shortstack{MM-IMDb\cite{arevalo2017mmimdb} \\ \\ (F1-Macro)}} 
    &\multirow{3}{*}{70\%} 
    &100\% &30\% & 100\% &30\% &35.13 &38.16 & \textbf{39.22} \\
    &&30\% &100\% & 30\% &100\% &37.73 &44.74 & \textbf{46.30} \\
    &&65\% &65\% & 65\% &65\% &36.26 &41.56 &\textbf{42.66} \\
    \hline     

    \multirow{3}{*}{\shortstack{Food101\cite{wang2015food101} \\ \\ (Accuracy)}} 
    &\multirow{3}{*}{70\%} 
    &100\% &30\% & 100\% &30\% &66.29 &72.57 &\textbf{74.53} \\
    &&30\% &100\% & 30\% &100\% &76.66 &86.05 &\textbf{86.18} \\
    &&65\% &65\% & 65\% &65\% &69.25 &78.09 &\textbf{79.08}  \\
    \hline     
   
    \multirow{3}{*}{\shortstack{Hateful Memes\cite{kiela2020hateful} \\ \\ (AUROC)}} 
    &\multirow{3}{*}{70\%} 
    &100\% &30\% & 100\% &30\% &60.78 &\textbf{62.17} &59.11 \\
    &&30\% &100\% & 30\% &100\% &61.64 &62.34 &\textbf{63.06} \\
    &&65\% &65\% & 65\% &65\% &62.48 &64.55 &\textbf{66.07} \\
    \end{tabular}
    }
    \vspace{-2mm}
\end{table*}

\begin{figure*}[!t]
    \centering
    \includegraphics[width=0.95\textwidth]{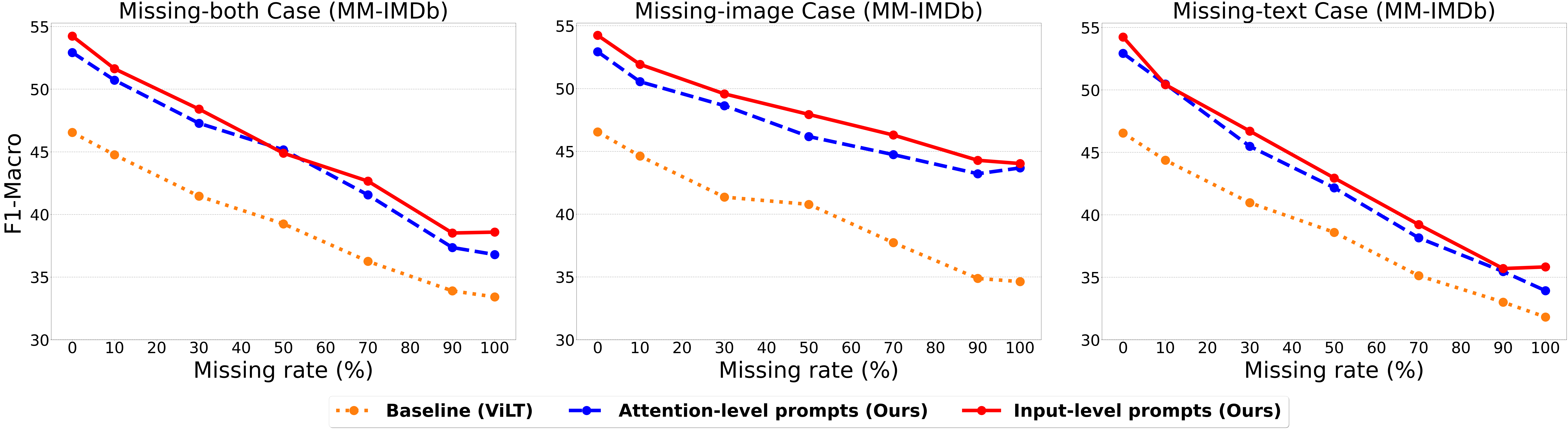}
    \caption{Quantitative results on the MM-IMDb dataset with different missing rates under different missing-modality scenarios. Each data point on the figure represents that training and testing are with the same $\eta\%$ missing rate.}
    \label{fig:ablation_missing_ratio_same_train_test}
    \vspace{-2mm}
\end{figure*}

\subsection{Implementation Details}
\vspace{-2mm}
\paragraph{Input.}
For the text modality, we use bert-base-uncased tokenizer to tokenize the text input. If the text is missing, we use an empty string as dummy input (i.e., $\tilde{x}^{m_1}$). The maximum length of text inputs varies according to the datasets: 1024 for MM-IMDb, 512 for UPMC Food-101, and 128 for Hateful Memes. For the image modality, we follow~\cite{kim2021vilt} to resize the shorter side of input images to 384 and limit the longer side to be under 640 while keeping the aspect ratio. Similar to \cite{dosovitskiy2020vit}, 
we decompose images into patches of size $32 \times 32$.
If the image is missing, we create an image with all pixel values equal to one as dummy input (i.e., $\tilde{x}^{m_2}$).

\vspace{-4mm}
\paragraph{Multimodal Backbone.} 
We adopt the pre-trained multimodal transformer ViLT~\cite{kim2021vilt} as our backbone since it is commonly used in various transformer-based methods for learning multimodal tasks. ViLT stems from Vision Transformers~\cite{dosovitskiy2020vit} and advances to process multimodal inputs with the tokenized texts and patched images. Without using modality-specific feature extractors, ViLT is pre-trained on several large vision-language datasets (e.g., MS-COCO~\cite{lin2014microsoft} and Visual Genome~\cite{krishna2017visual}) via objectives such as Image Text Matching (ITM) and Masked Language Modeling (MLM).

\vspace{-4mm}
\paragraph{Model Training Details.} 
We freeze all the parameters of the ViLT backbone to avoid heavy finetuning and only train the learnable prompts and parameters corresponding to downstream tasks (i.e., the pooler and task-specific classifier). The length $L_p$ of learnable prompts is set to 16 by default, and the indices of MSA layers to attach prompts start from 0 and end at 5 (i.e., maximum 6 MSA layers are prompted).
We use the AdamW optimizer\cite{loshchilov2017decoupled} in all
experiments. The base learning rate is $1\times 10^{-2}$ and weight decay is $2\times 10^{-2}$. The learning rate is warmed up for 10\% of the total training steps and then is decayed linearly to zero. 

\vspace{-4mm}
\paragraph{Setting of Missing Modality.} 
We focus on the more general scenario where the missing modality can occur in both training and testing phases, where each modality for each data sample has chances to be lost. We define the missing rate $\eta\%$ as the proportion of modality-incomplete data to the entire dataset. In the vision and language tasks, there are three possible cases of missing modality: missing-text, missing-image, and missing-both. For the first two cases, missing-text (missing-image) with missing rate $\eta\%$ indicates that there are $\eta\%$ image-only (text-only) data and $(1-\eta)\%$ complete data. For the missing-both case, there are $\frac{\eta}{2}\%$ text-only data, $\frac{\eta}{2}\%$ image-only data, and $(1-\eta)\%$ complete data. Such partition can be extended to the tasks of $M$ modality: having $(\frac{\eta}{M^2-2})\%$ modality-incomplete data for each missing case and $(1-\eta)\%$ complete data. In our experiments, the missing rate $\eta\%$ is set to 70\% by default.

\begin{figure*}[!t]
    \centering
    \Huge
    \resizebox{1\linewidth}!{
    \begin{tabular}{ccc}
    \begin{tabular}{c}\includegraphics[width=0.95\textwidth]{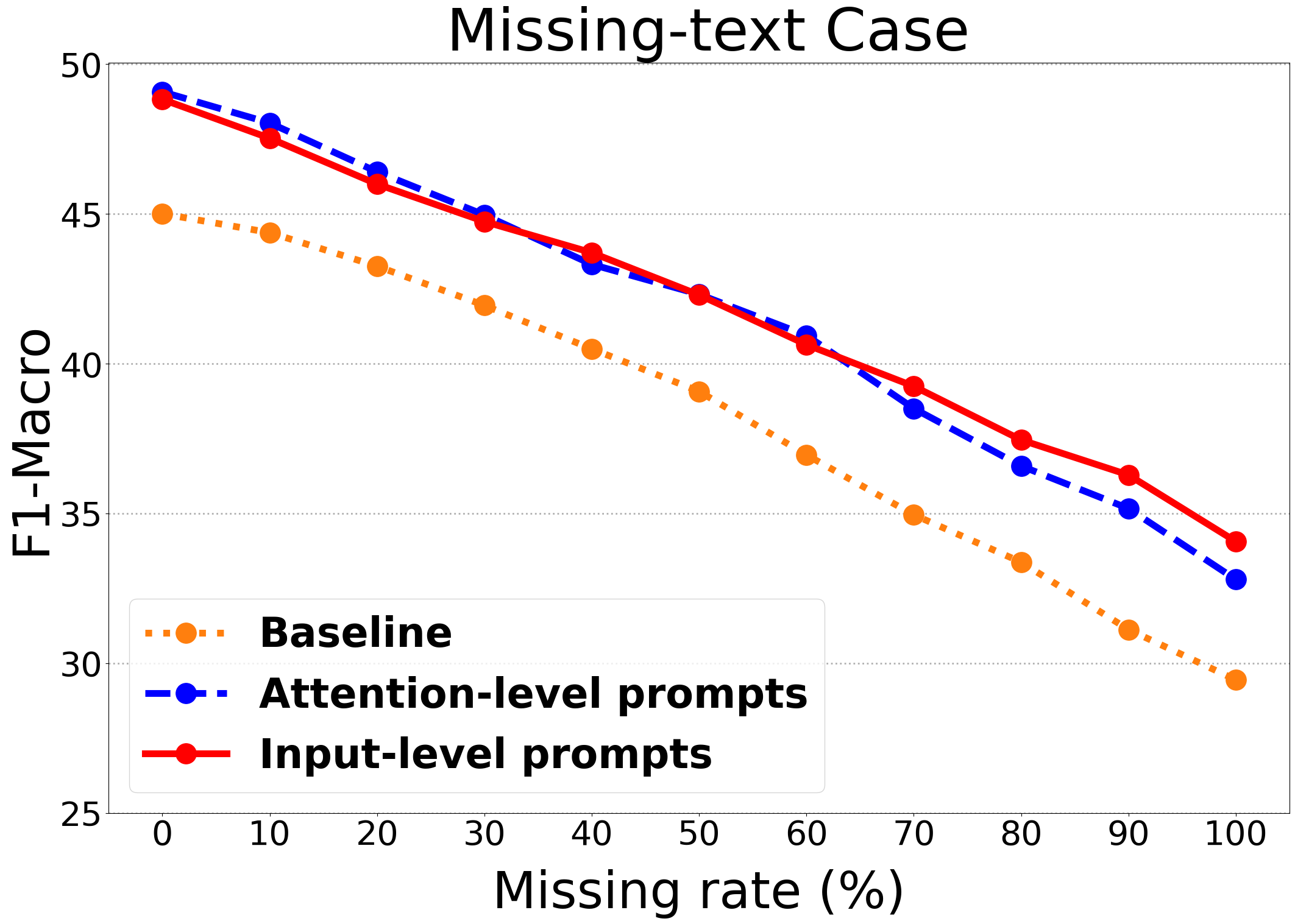}\end{tabular}
    &\begin{tabular}{c}\includegraphics[width=0.95\textwidth]{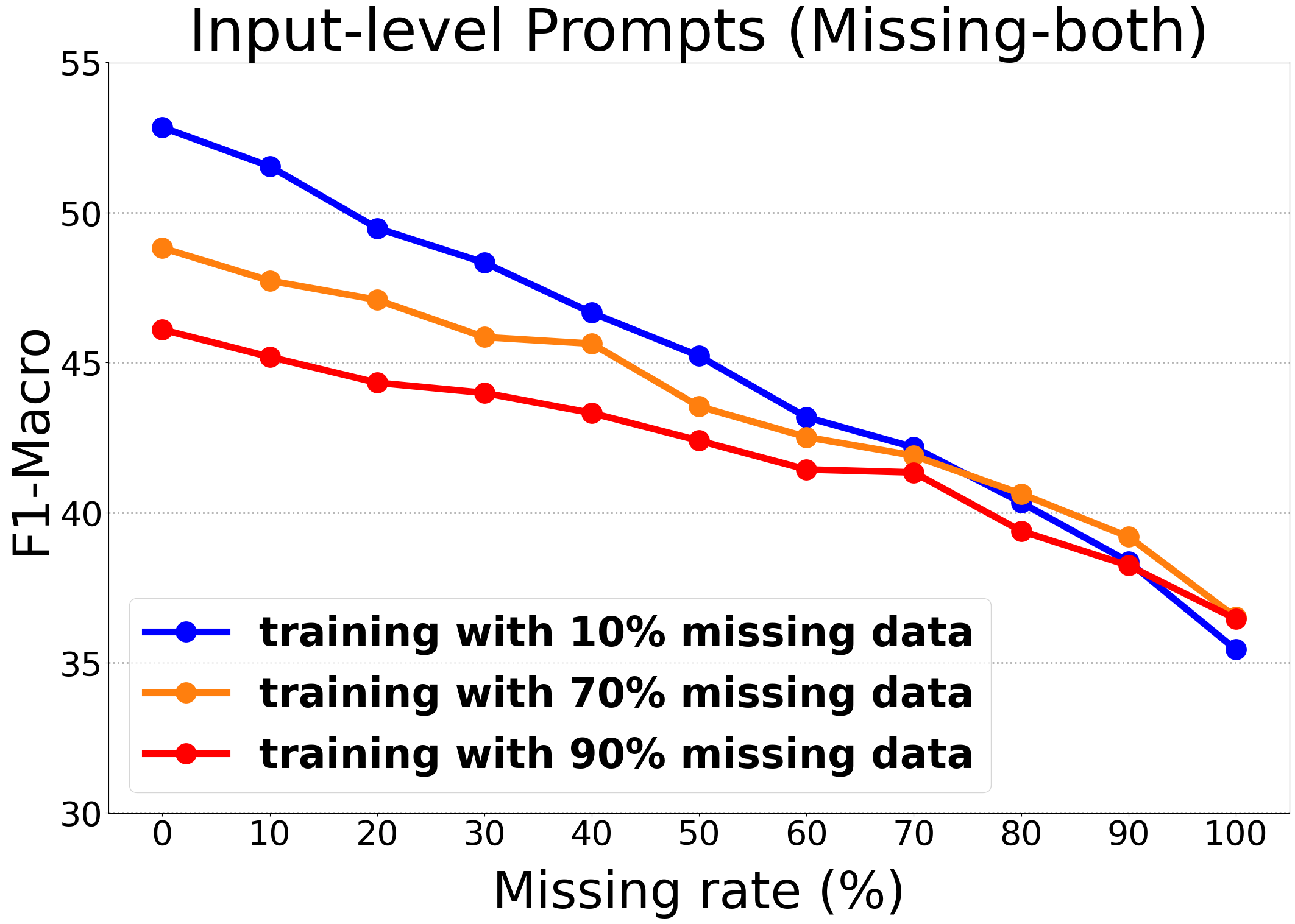}\end{tabular} 
    &\begin{tabular}{c}\includegraphics[width=0.95\textwidth]{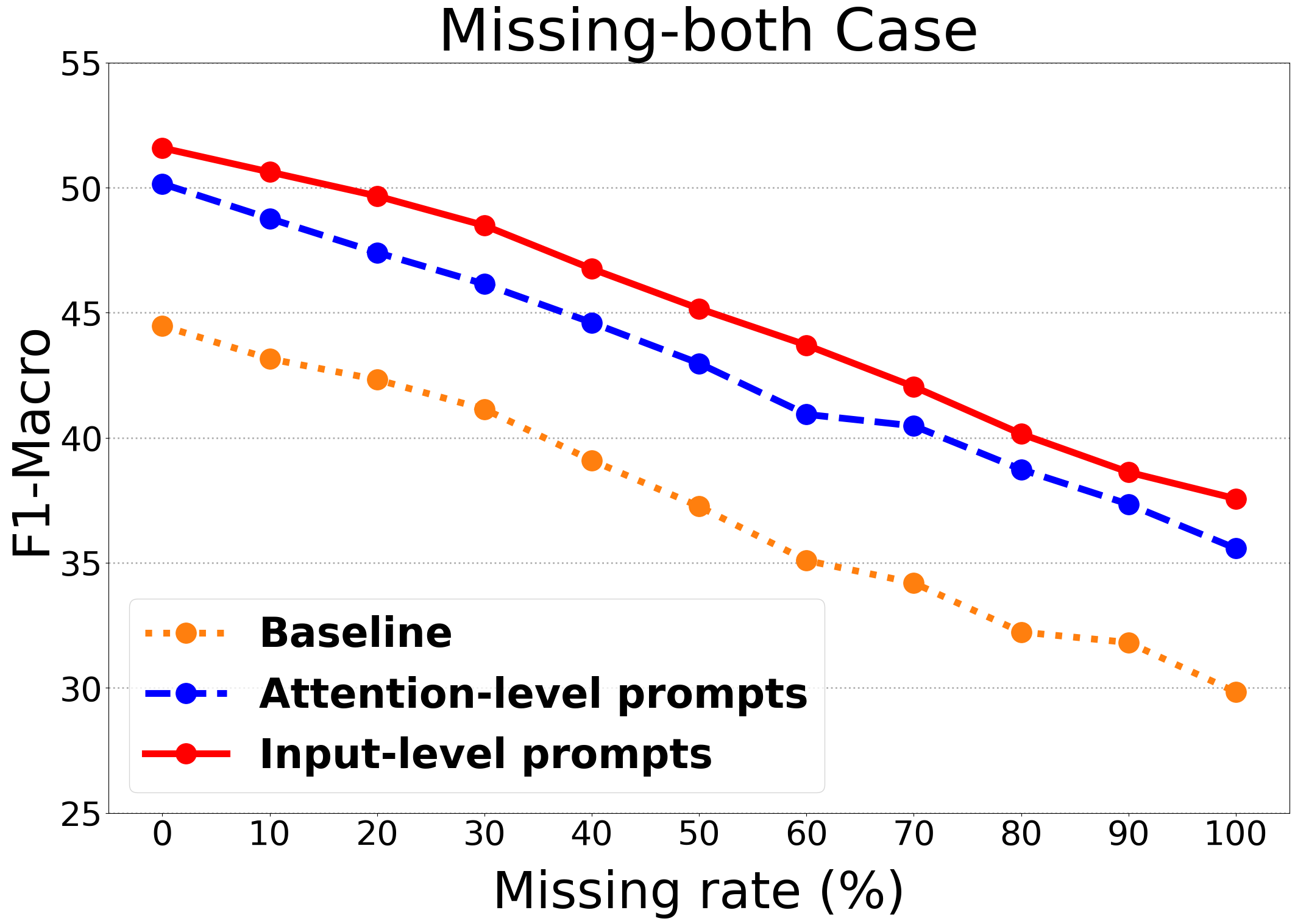}\end{tabular}\\
    (a) \textcolor{blue}{Train}: Missing-both 70$\%$; \textcolor{red}{Test}: Missing-text  
    &(b) \textcolor{blue}{Train}: Missing-both; \textcolor{red}{Test}: Missing-both 
    &(c) \textcolor{blue}{Train}: Modality-complete; \textcolor{red}{Test}: Missing-both
    \end{tabular}
    }
    \caption{Ablation study on robustness to the testing missing rate in different scenarios on MM-IMDb. (a) All models are trained on missing-both case with 70\% missing rate, and evaluated on missing-text case with different missing rates. (b) Input-level prompts are trained on missing-both cases with 10\%, 70\%, and 90\% missing rate, which represents more modality-complete data, balanced data, and less modality-complete data, respectively. Evaluation is on missing-both case with different missing rates. (c) All models are trained with modality-complete data, where each data pair can be randomly assigned with different missing modality at different training epochs (i.e., text-only, image-only, and modality-complete) to account for the possible missing modalities in the testing time. Evaluation is on missing-both case with different missing rates.}
    \label{fig:ablations}
    \vspace{-2mm}
\end{figure*}

\subsection{Main Results}
We focus on studying the robustness of multimodal transformers against general incompleteness in multimodal data, without the requirement to finetune the entire heavy model.
The main baseline to compare with is the model that only trains the pooler and the task-specific classifier, where the improved performance brought by our proposed method with respect to such baseline directly reflects the benefit from our missing-aware prompts.

In Table~\ref{table:qualitative}, we show quantitative results on three classification tasks with 1) various missing-modality cases, and 2) two prompt learning designs.
First, we find that our attention-level prompting consistently improves the baseline by a large margin in all the scenarios, showing that our missing-aware prompts, without entire model finetuning, are able to tackle general missing-modality cases and provide a good instruction for model prediction. 
Next, 
we show that input-level prompting further improves the performance in most settings, except for one case on the Hateful Memes dataset, in which this sensitivity analysis will be discussed in the following section.

In Figure~\ref{fig:ablation_missing_ratio_same_train_test}, we present more results from a wider range of missing rate on MM-IMDb. Similar trends can be concluded, where the input-level prompting shows favorable performance compared to the other two methods.
More qualitative results, in which shows similar trends, are provided in the supplementary material due to limited space.

\label{sec:performance_sensitivity}
\vspace{-4mm}
\paragraph{Performance Sensitivity.}
As introduced in Section~\ref{sec:method_prompt_func}, different configurations of prompts have distinct behaviors to learn the instruction for pre-trained models.
Input-level prompting can learn more effective instructions for each prompting layer, with the information of their inherited prompt tokens from previous layers.
However, the performance may become sensitive, depending on the prompt length and the input length of different datasets.
For instance, the Hateful Memes dataset has shorter text sequences (i.e., 128) compared to other datasets, in which prompts may become more influential to the final performance if the prompt length is too long.
This may cause ambiguity for the model to learn task-specific features, e.g., input-level prompting performs slightly worse on Hateful Memes, especially for the text-missing case in Table~\ref{table:qualitative}.

In contrast, attention-level prompts are attached to the key and value of each prompted MSA layer, and hence each prompt is only responsible for the instruction of its corresponding layer.
While input-level prompting reaches the best performance in most cases, attention-level prompting can be more stable and less sensitive to different datasets.
This observation indicates the trade-off in terms of model performance and stability for different prompt designs.

\vspace{-4mm}
\paragraph{Efficiency on Parameters.} In comparison with finetuning the entire model which needs to update 113M parameters of the ViLT, resulting in 46.45 F1-Macro score on MM-IMDb for the missing-both case, our proposed method keeps the pre-trained ViLT frozen and only needs to train the missing-aware prompts of 221K parameters (i.e., merely 0.2\% compared to the entire model), while still achieving competitive performance of 42.66 in F1-Macro (see Table~\ref{table:qualitative}). Note that, the number of parameters for the task classifier is not taken into consideration as it is necessary for learning multimodal tasks. Particularly, once scaling up to the huge models with billions of parameters, our proposed missing-aware prompting would be more preferable and applicable for multimodal downstream tasks with missing modalities, with a better balance between computational cost and performance.

\subsection{Ablation Study}
\vspace{-2mm}
\paragraph{Robustness to different missing rates.} 

We conduct further experiments to analyze the robustness of our proposed method against different missing-modality rates between training and testing phases.
We first evaluate the models that are trained on the missing-both case with missing rate $70\%$, and test with different missing-text situations. As shown in Figure~\ref{fig:ablations}(a), we find that both attention-level and input-level promptings are robust to different missing rates in testing time, comparing with the baseline.
Moreover, we observe the trend that attention-level prompting performs better than the input-level one when testing with a missing rate lower than 30\%, but gradually become less effective when the missing rate goes higher than 30\%.
Such a trend indicates that attention-level prompting learns better on modality-complete data, while input-level prompting is more robust to modality-incomplete data.

Moreover, in Figure~\ref{fig:ablations}(b), we examine three input-level prompting models trained with missing-both rates 10\%, 70\%, and 90\% to represent the degree of missing situations during training: more modality-complete data (90\% complete), balanced data (30\% complete, 35\% only-text, and 35\% only-image), and less modality-complete data (10\% complete) respectively.
We find that when training with more modality-complete (i.e., low missing rate) data, the performance is much higher when also testing with low missing rate.
Interestingly, when testing with a high missing rate, the model trained with 90\% incomplete data performs competitively even with model trained with much more complete data. This shows the robustness of our prompt design to handle different missing-modality cases.

\vspace{-4mm}
\paragraph{Results with complete training data.} 
In this paper, we assume that it is not guaranteed to collect entire modality-complete data due to privacy and budget limits in the real-world scenario. However, there still exists publicly available modality-complete datasets which are well annotated for training. Therefore, we conduct experiments with a new baseline and our methods by training with modality-complete data.
To account for the missing-modality case during testing, we randomly select data with different modality-missing cases for each data pair (i.e., text-only, image-only, or complete). Note that, different from other experimental settings introduced in the paper, here one data pair can be in various missing-modality cases at different training epochs.
The results are shown in Figure~\ref{fig:ablations}(c). Similarly, we find that our method consistently improves the baseline under different missing rates.

\begin{figure}
    \centering
    \includegraphics[width=1\textwidth]{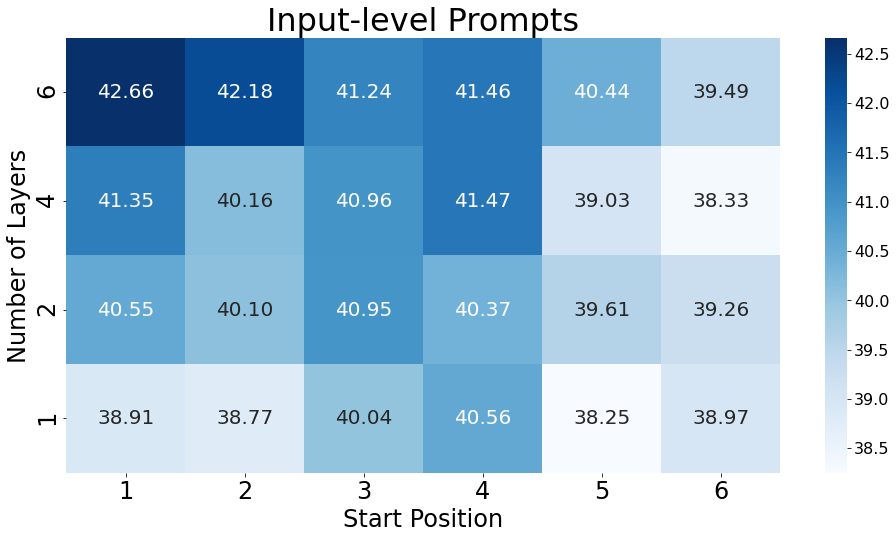}
    \caption{Ablation study on the location of prompting layers for input-level prompts.
    }
    \label{fig:ablation_grid_search}
    \vspace{-3mm}
\end{figure}

\vspace{-4mm}
\paragraph{The effect of selected layers ($start, end$).} 
\label{sec:prompt_position}
We conduct experiments to analyze the effect of locations to attach prompting layers in Figure~\ref{fig:ablation_grid_search}. We observe that the performance increases intuitively as the number of layers increases, while a more critical factor is which layer to start attaching prompts, i.e., the earlier layer the better.
This indicates that early layers with prompts influence model predictions more.
One reason can be: multimodal inputs are fused from the beginning of the transformer, in which the degree of fusion increases when the layer is deeper (i.e., the characteristics of each modality remain more distinct in earlier layers). Therefore, it is more effective for early layers to be guided by the instruction of missing-aware prompts, before each modality loses its distinct characteristics.

\begin{figure}
    \centering
    \includegraphics[width=0.9\textwidth]{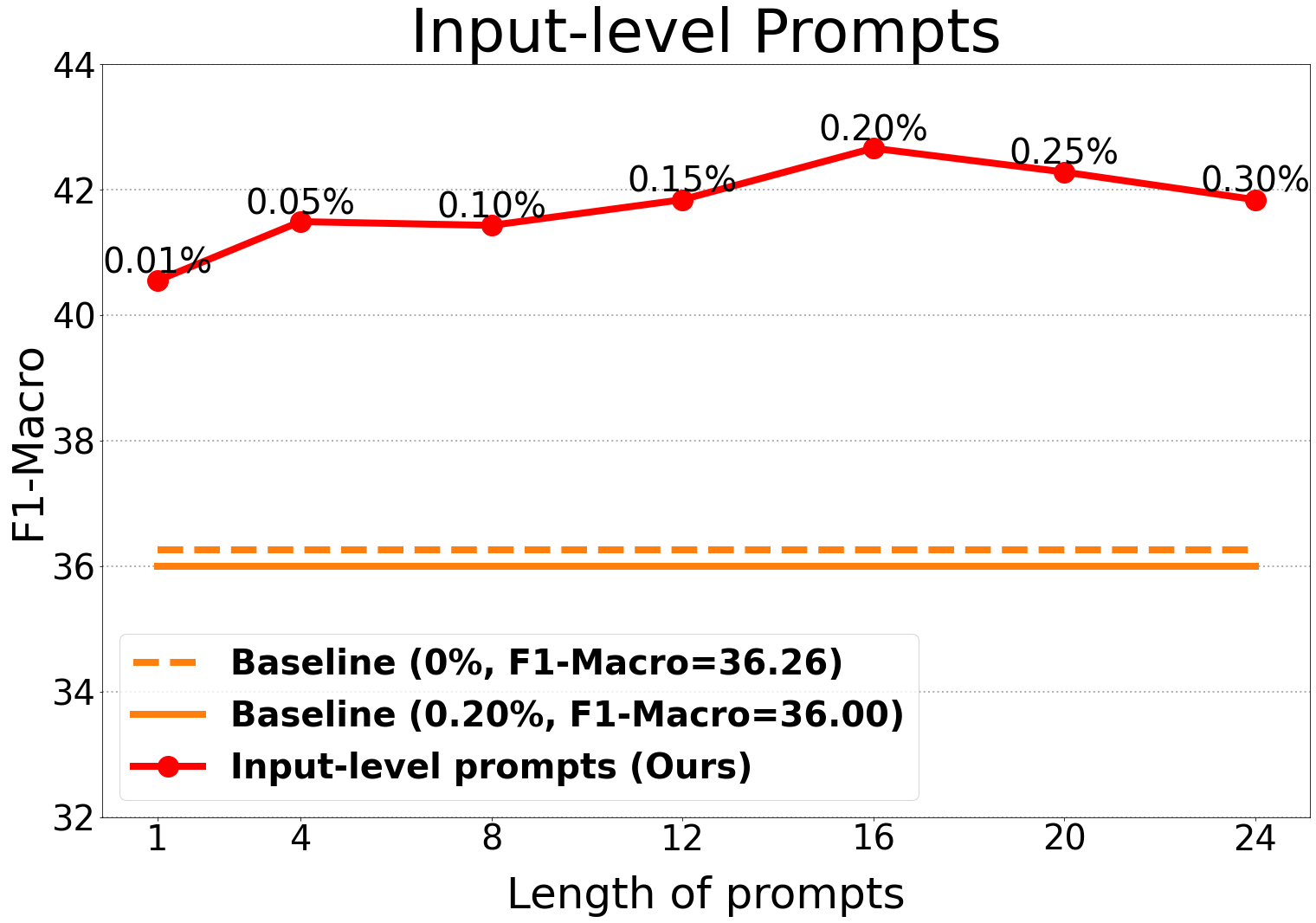}
    \caption{Ablation study on different length $L_P$ of prompts for input-level prompts. The numbers above the red points are the proportion of parameters in prompts, compared to the entire model. We further conduct the new baseline with additional parameters with the same proportion (e.g., 0.2\%) of the prompt size, denoted as the orange solid line.}
    \label{fig:ablation_prompt_length}
    \vspace{-3mm}
\end{figure}

\vspace{-4mm}
\paragraph{The effect of prompt length $L_p$.} 
\label{sec:prompt_length}
We study the influence of prompt length in Figure~\ref{fig:ablation_prompt_length}. Intuitively, the performance is improved as the prompt length becomes longer. However, both prompting methods reach the best performance with the prompt length equal to 16, showing that the length should be balanced. 
In addition, to validate the efficiency of our method, we calculate the proportion of prompt parameters to parameters of the entire pre-trained model (numbers above the red data points in Figure~\ref{fig:ablation_prompt_length}.
Our prompt-based method only requires an additional 0.2\% parameters but improves baselines by a large margin. Even with fewer parameters (i.e., reducing the prompt length to 1), the performance is still competitive.
We also conduct a new baseline with additional parameters in the task classifier of the same proportion, i.e., 0.2\% of the entire model parameters, but it does not show clear improvement. It demonstrates the efficacy of the proposed method in learning multimodal tasks to handle missing-modality, only with very few trainable parameters.

\vspace{-4mm}
\paragraph{Limitations and future works.} 
Though our modality-missing-aware prompting can largely increase the robustness of the tuned backbone models, it does not recover the missing information from the multimodal input. We expect the cross-modal generative modeling can help further boost the performance by generating missing information.
Besides, when facing the high-modality scenario, there could be quadratic growth on number of prompts. To tackle this issue, we expect to adopt the prompt pool concept in the recent L2P~\cite{wang2022learning}
work, where the prompting mechanism can query from a fixed number of prompts in a designated pool to avoid the quadratic growth.

\section{Conclusions}
\noindent We tackle two major challenges in multimodal learning: 1) a general scenario for missing modality that occurs either during training or testing, and 2) heavy computational requirement for training transformers.
As a simple yet effective approach, we propose a missing-aware prompting method which is easy to plug in the transformer-like multimodal model to alleviate the performance drop caused by missing modality, while also not requiring heavy model finetuning.
We further explore the configuration of prompts and show the robustness to the missing modalities during various scenarios. Extensive experiments and ablation studies demonstrate the effectiveness of our approach.

\noindent \textbf{Acknowledgement}~
This work is supported by NSTC (National Science and Technology Council, Taiwan) 111-2628-EA49-018-MY4 and 111-2636-E-A49-003. 

{\small
\bibliographystyle{ieee_fullname}
\bibliography{egbib}
}

\end{document}